\documentclass{article} 
\usepackage{nips14submit_e,times}
\usepackage{hyperref}
\usepackage{url}
\usepackage{natbib}
\usepackage{amsmath}
\usepackage{amsthm}
\usepackage{amssymb}
\usepackage{graphicx}
\usepackage{xspace}
\usepackage{tabularx}
\usepackage{multirow}
\usepackage{wrapfig}

\newcommand{\bmx}[0]{\begin{bmatrix}}
\newcommand{\emx}[0]{\end{bmatrix}}

\newcommand{\vect}[1]{\mathbf{#1}}
\newcommand{\vects}[1]{\boldsymbol{#1}}

\newcommand{\vc}[0]{\vect{c}}
\newcommand{\vh}[0]{\vect{h}}

\newcommand{\vx}[0]{\vect{x}}
\newcommand{\vr}[0]{\vect{r}}

\newcommand{\vy}[0]{\vect{y}}

\newcommand{\TT}[0]{\vects{\theta}}

\newcommand{\specialcell}[2][c]{%
  \begin{tabular}[#1]{@{}c@{}}#2\end{tabular}}

\nipsfinalcopy 

\title{Empirical Evaluation of \\ Gated Recurrent Neural Networks \\ on Sequence Modeling}

\author{
Junyoung Chung ~ ~ ~ Caglar Gulcehre ~ ~ ~
KyungHyun Cho\\
Universit\'{e} de Montr\'{e}al
\And
Yoshua Bengio\\
Universit\'{e} de Montr\'{e}al \\
CIFAR Senior Fellow
}

\begin{document}

\maketitle
\begin{abstract}
    In this paper we compare different types of recurrent units in recurrent
neural networks (RNNs). Especially, we focus on more sophisticated units that
implement a gating mechanism, such as a long short-term memory (LSTM) unit
and a recently proposed gated recurrent unit (GRU). We evaluate these
recurrent units on the tasks of polyphonic music modeling and speech signal
modeling. Our experiments revealed that these advanced recurrent units are
indeed better than more traditional recurrent units such as $\tanh$ units.
Also, we found GRU to be comparable to LSTM.
\end{abstract}

\section{Introduction}

Recurrent neural networks have recently shown promising results in many machine
learning tasks, especially when input and/or output are of variable
length~\citep[see, e.g.,][]{Graves-book2012}. More recently, \citet{Sutskever-et-al-arxiv2014}
and \citet{Bahdanau-et-al-arxiv2014} reported that recurrent neural networks are
able to perform as well as the existing, well-developed systems on a challenging
task of machine translation.

One interesting observation, we make from these recent successes is that almost
none of these successes were achieved with a vanilla recurrent neural network. Rather, it was a
recurrent neural network with sophisticated recurrent hidden units, such as long
short-term memory units~\citep{Hochreiter+Schmidhuber-1997}, that was used in
those successful applications.

Among those sophisticated recurrent units, in this paper, we are interested in
evaluating two closely related variants. One is a long short-term memory (LSTM)
unit, and the other is a gated recurrent unit (GRU) proposed more recently by
\citet{cho2014properties}. It is well established in the field that the LSTM
unit works well on sequence-based tasks with long-term dependencies, but the
latter has only recently been introduced and used in the context of machine
translation.

In this paper, we evaluate these two units and a more traditional
$\tanh$ unit on the task of sequence modeling. We consider three
polyphonic music datasets~\citep[see, e.g.,][]{Boulanger-et-al-ICML2012} as well as two
internal datasets provided by Ubisoft in which each sample is a raw speech
representation.

Based on our experiments, we concluded that by using fixed number of parameters for all models
on some datasets GRU, can outperform LSTM units both in terms of convergence in CPU time and
in terms of parameter updates and generalization.

\section{Background: Recurrent Neural Network}

A recurrent neural network (RNN) is an extension of a conventional feedforward neural
network, which is able to handle a variable-length sequence input. The RNN
handles the variable-length sequence by having a recurrent hidden state whose
activation at each time is dependent on that of the previous time.

More formally, given a sequence $\vx=\left( \vx_1, \vx_2, \cdots, \vx_{\scriptscriptstyle{T}} \right)$, the
RNN updates its recurrent hidden state $h_t$ by
\begin{align}
    \label{eq:rnn_hidden}
    \vh_t =& \begin{cases}
        0, & t = 0\\
        \phi\left(\vh_{t-1}, \vx_{t}\right), & \mbox{otherwise}
    \end{cases}
\end{align}
where $\phi$ is a nonlinear function such as composition of a logistic sigmoid with an affine transformation.
Optionally, the RNN may have an output $\vy=\left(y_1, y_2, \dots, y_{\scriptscriptstyle{T}}\right)$ which
may again be of variable length.

Traditionally, the update of the recurrent hidden state in Eq.~\eqref{eq:rnn_hidden} is implemented as

\begin{align}
    \label{eq:rnn_trad}
    \vh_t = g\left( W \vx_t + U \vh_{t-1} \right),
\end{align}
where $g$ is a smooth, bounded function such as a logistic sigmoid function or a
hyperbolic tangent function.

A generative RNN outputs a probability distribution over the next element of the sequence,
given its current state $\vh_t$, and this generative model can capture a distribution
over sequences of variable length by using a special output symbol to represent
the end of the sequence. The sequence probability can be decomposed into

\begin{align}
\label{eq:seq_model}
    p(x_1, \dots, x_{\scriptscriptstyle{T}}) = p(x_1) p(x_2\mid x_1) p(x_3 \mid x_1, x_2) \cdots
    p(x_{\scriptscriptstyle{T}} \mid x_1, \dots, x_{\scriptscriptstyle{T-1}}),
\end{align}

where the last element is a special end-of-sequence value. We model each conditional probability distribution with

\begin{align*}
    p(x_t \mid x_1, \dots, x_{t-1}) =& g(h_t),
\end{align*}
where $h_t$ is from Eq.~\eqref{eq:rnn_hidden}. Such generative RNNs are the subject
of this paper.

Unfortunately, it has been observed by, e.g., \citet{Bengio-trnn94} that it is
difficult to train RNNs to capture long-term dependencies because the gradients
tend to either vanish (most of the time) or explode (rarely, but with severe effects).
This makes gradient-based optimization method struggle, not just because of the
variations in gradient magnitudes but because of the effect of long-term dependencies
is hidden (being exponentially smaller with respect to sequence length) by the
effect of short-term dependencies.
There have been two dominant approaches by which many researchers have tried to reduce the negative
impacts of this issue. One such approach is to devise a better learning algorithm
than a simple stochastic gradient descent~\citep[see, e.g.,][]{Bengio-et-al-ICASSP-2013,Pascanu-et-al-ICML2013,Martens+Sutskever-ICML2011},
for example using the very simple {\em clipped gradient}, by which the norm of the
gradient vector is clipped, or using second-order methods which may be less sensitive
to the issue if the second derivatives follow the same growth pattern as the first derivatives
(which is not guaranteed to be the case).

The other approach, in which we are more interested in this paper, is to design
a more sophisticated activation function than a usual activation function,
consisting of affine transformation followed by a simple element-wise
nonlinearity by using gating units. The earliest attempt in this direction
resulted in an activation function, or a recurrent unit, called a long short-term memory (LSTM)
unit~\citep{Hochreiter+Schmidhuber-1997}. More recently, another type of
recurrent unit, to which we refer as a gated recurrent unit (GRU), was proposed
by \citet{cho2014properties}. RNNs employing either of these recurrent units
have been shown to perform well in tasks that require capturing long-term
dependencies. Those tasks include, but are not limited to, speech
recognition~\citep[see, e.g.,][]{Graves-et-al-ICASSP2013} and machine
translation~\citep[see, e.g.,][]{Sutskever-et-al-arxiv2014,Bahdanau-et-al-arxiv2014}.


\section{Gated Recurrent Neural Networks}

In this paper, we are interested in evaluating the performance of those recently
proposed recurrent units (LSTM unit and GRU) on sequence modeling.
Before the empirical evaluation, we first describe each of those recurrent units
in this section.

\begin{figure}[t]
    \centering
    \begin{minipage}[b]{0.4\textwidth}
        \centering
        \includegraphics[height=0.18\textheight]{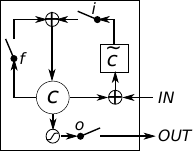}
    \end{minipage}
    \hspace{5mm}
    \begin{minipage}[b]{0.4\textwidth}
        \centering
        \includegraphics[height=0.16\textheight]{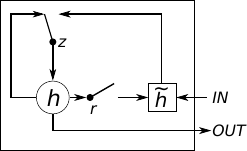}
    \end{minipage}

    \vspace{2mm}
    \begin{minipage}{0.4\textwidth}
        \centering
        (a) Long Short-Term Memory
    \end{minipage}
    \hspace{5mm}
    \begin{minipage}{0.4\textwidth}
        \centering
        (b) Gated Recurrent Unit
    \end{minipage}

    \caption{
        Illustration of (a) LSTM and (b) gated recurrent units. (a) $i$, $f$ and
        $o$ are the input, forget and output gates, respectively. $c$ and
        $\tilde{c}$ denote the memory cell and the new memory cell content. (b)
        $r$ and $z$ are the reset and update gates, and $h$ and $\tilde{h}$ are
        the activation and the candidate activation.
    }
    \label{fig:gated_units}
\end{figure}

\subsection{Long Short-Term Memory Unit}
\label{sec:lstm}

The Long Short-Term Memory (LSTM) unit was initially proposed by
\citet{Hochreiter+Schmidhuber-1997}. Since then, a number of minor modifications
to the original LSTM unit have been made. We follow the implementation of LSTM as used in
\citet{graves2013generating}.

Unlike to the recurrent unit which simply computes a weighted sum of the input signal and
applies a nonlinear function, each $j$-th LSTM unit maintains a memory $c_t^j$ at
time $t$. The output $h_t^j$, or the activation, of the LSTM unit is then
\begin{align*}
    h_t^j = o_t^j \tanh \left( c_t^j \right),
\end{align*}
where $o_t^j$ is an {\it output gate} that modulates the amount of memory content
exposure. The output gate is computed by
\begin{align*}
    o_t^j = \sigma\left(
    W_o \vx_t + U_o \vh_{t-1} + V_o \vc_t
    \right)^j,
\end{align*}
where $\sigma$ is a logistic sigmoid function. $V_o$ is a diagonal matrix.

The memory cell $c_t^j$ is updated by partially forgetting the existing memory and adding
a new memory content $\tilde{c}_t^j$ :
\begin{align}
    \label{eq:lstm_memory_up}
    c_t^j = f_t^j c_{t-1}^j + i_t^j \tilde{c}_t^j,
\end{align}
where the new memory content is
\begin{align*}
    \tilde{c}_t^j = \tanh\left( W_c \vx_t + U_c \vh_{t-1}\right)^j.
\end{align*}

The extent to which the existing memory is forgotten is modulated by a {\it
forget gate} $f_t^j$, and the degree to which the new memory content is added to
the memory cell is modulated by an {\it input gate} $i_t^j$. Gates are computed by
\begin{align*}
    f_t^j =& \sigma\left( W_f \vx_t + U_f \vh_{t-1} + V_f \vc_{t-1} \right)^j, \\
    i_t^j =& \sigma\left( W_i \vx_t + U_i \vh_{t-1} + V_i \vc_{t-1} \right)^j.
\end{align*}
Note that $V_f$ and $V_i$ are diagonal matrices.

Unlike to the traditional recurrent unit which overwrites its content at each
time-step (see Eq.~\eqref{eq:rnn_trad}), an LSTM unit is able to decide whether
to keep the existing memory via the introduced gates. Intuitively, if the LSTM
unit detects an important feature from an input sequence at early stage, it
easily carries this information (the existence of the feature) over a long
distance, hence, capturing potential long-distance dependencies.

See Fig.~\ref{fig:gated_units}~(a) for the graphical illustration.

\subsection{Gated Recurrent Unit}
\label{sec:gru}

A gated recurrent unit (GRU) was proposed by \cite{cho2014properties} to make
each recurrent unit to adaptively capture dependencies of different time scales.
Similarly to the LSTM unit, the GRU has gating units that modulate the flow of information
inside the unit, however, without having a separate memory cells.

The activation $h_t^j$ of the GRU at time $t$ is a linear interpolation between
the previous activation $h_{t-1}^j$ and the candidate activation $\tilde{h}_t^j$:
\begin{align}
    \label{eq:gru_memory_up}
    h_t^j = (1 - z_t^j) h_{t-1}^j + z_t^j \tilde{h}_t^j,
\end{align}
where an {\it update gate} $z_t^j$ decides how much the unit updates its activation,
or content. The update gate is computed by
\begin{align*}
    z_t^j = \sigma\left( W_z \vx_t + U_z \vh_{t-1} \right)^j.
\end{align*}

This procedure of taking a linear sum between the existing state and the newly
computed state is similar to the LSTM unit. The GRU, however, does not have any
mechanism to control the degree to which its state is exposed, but exposes the
whole state each time.

The candidate activation $\tilde{h}_t^j$ is computed similarly to that of the traditional
recurrent unit (see Eq.~\eqref{eq:rnn_trad}) and as in \citep{Bahdanau-et-al-arxiv2014},

\begin{align*}
   \tilde{h}_t^j = \tanh\left( W \vx_t + U \left(  \vr_t \odot \vh_{t-1}\right) \right)^j,
\end{align*}
where $\vr_t$ is a set of reset gates and $\odot$ is an element-wise
multiplication.
\footnote{
    Note that we use the reset gate in a slightly different way from the
original GRU proposed in \cite{cho2014properties}. Originally, the
candidate activation was computed by
\begin{align*}
    \tilde{h}_t^j = \tanh\left( W \vx_t + \vr_t\odot\left( U \vh_{t-1}\right) \right)^j,
\end{align*}
where $r_t^j$ is a {\it reset gate}.
We found in our preliminary experiments that both of these
    formulations performed as well as each other.
}
When off ($r_t^j$ close to $0$), the reset gate effectively makes the
unit act as if it is reading the first symbol of an input sequence, allowing it
to {\it forget} the previously computed state.

The reset gate $r_t^j$ is computed similarly to the update gate:
\begin{align*}
    r_t^j = \sigma\left( W_r \vx_t + U_r \vh_{t-1} \right)^j.
\end{align*}
See Fig.~\ref{fig:gated_units}~(b) for the graphical illustration of the GRU.

\subsection{Discussion}

It is easy to notice similarities between the LSTM unit and the GRU
from Fig.~\ref{fig:gated_units}.

The most prominent feature shared between these units is the additive component of their
update from $t$ to $t+1$, which is lacking in the traditional recurrent unit. The traditional recurrent
unit always replaces the activation, or the content of a unit with a new value
computed from the current input and the previous hidden state. On the other
hand, both LSTM unit and GRU keep the existing content and add the new content
on top of it (see Eqs.~\eqref{eq:lstm_memory_up}~and~\eqref{eq:gru_memory_up}).

This additive nature has two advantages. First, it is easy for each unit to
remember the existence of a specific feature in the input stream for a long series of steps.
Any important feature, decided by either the forget gate of the LSTM unit or the
update gate of the GRU, will not be overwritten but be maintained as it is.

Second, and perhaps more importantly, this addition effectively creates shortcut
paths that bypass multiple temporal steps. These shortcuts allow the error to be
back-propagated easily without too quickly vanishing (if the gating unit is nearly
saturated at $1$) as a result of passing through multiple, bounded nonlinearities,
thus reducing the difficulty due to vanishing gradients~\citep{Hochreiter91,Bengio-trnn94}.

These two units however have a number of differences as well. One feature of the
LSTM unit that is missing from the GRU is the controlled exposure of the memory
content. In the LSTM unit, the amount of the memory content that is seen, or
used by other units in the network is controlled by the output gate. On
the other hand the GRU exposes its full content without any control.

Another difference is in the location of the input gate, or the corresponding
reset gate. The LSTM unit computes the new memory content without any separate
control of the amount of information flowing from the previous time step.
Rather, the LSTM unit controls the amount of the new memory content being added
to the memory cell {\it independently} from the forget gate. On the other hand,
the GRU controls the information flow from the previous activation when
computing the new, candidate activation, but does not independently control the
amount of the candidate activation being added (the control is tied via the
update gate).

From these similarities and differences alone, it is difficult to conclude which
types of gating units would perform better in general. Although \citet{Bahdanau-et-al-arxiv2014}
reported that these two units performed comparably to each other according to
their preliminary experiments on machine translation, it is unclear whether this
applies as well to tasks other than machine translation. This motivates us to
conduct more thorough empirical comparison between the LSTM unit and the GRU in
this paper.

\section{Experiments Setting}

\subsection{Tasks and Datasets}

We compare the LSTM unit, GRU and $\tanh$ unit in the task of sequence modeling.
Sequence modeling aims at learning a probability distribution over sequences, as
in Eq.~\eqref{eq:seq_model}, by maximizing the log-likelihood of a model given a
set of training sequences:
\begin{align*}
\max_{\TT} \frac{1}{N} \sum_{n=1}^N \sum_{t=1}^{T_n} \log p\left(x_t^n
        \mid x_1^n, \dots, x_{t-1}^n; \TT \right),
\end{align*}
where $\TT$ is a set of model parameters. More specifically, we evaluate these
units in the tasks of polyphonic music modeling and speech signal modeling.

For the polyphonic music modeling, we use three polyphonic music
datasets from \citep{Boulanger-et-al-ICML2012}: Nottingham, JSB
Chorales, MuseData and Piano-midi.  These datasets contain
sequences of which each symbol is respectively a $93$-, $96$-, $105$-, and
$108$-dimensional binary vector. We use logistic sigmoid function
as output units.

We use two internal datasets provided by Ubisoft\footnote{
    \url{http://www.ubi.com/}
} for speech signal modeling. Each sequence is an one-dimensional
raw audio signal, and at each time step, we design a recurrent
neural network to look at $20$ consecutive samples to predict the
following $10$ consecutive samples. We have used two different
versions of the dataset: One with sequences of length $500$
(Ubisoft A) and the other with sequences of length $8,000$ (Ubisoft B).
Ubisoft A and Ubisoft B have $7,230$ and $800$ sequences each.
We use mixture of Gaussians with 20 components as output layer.
\footnote{Our implementation is available at \url{https://github.com/jych/librnn.git}}

\subsection{Models}

For each task, we train three different recurrent neural
networks, each having either LSTM units (LSTM-RNN, see
Sec.~\ref{sec:lstm}), GRUs (GRU-RNN, see Sec.~\ref{sec:gru}) or $\tanh$
units ($\tanh$-RNN, see Eq.~\eqref{eq:rnn_trad}). As the primary objective of
these experiments is to compare all three units fairly, we choose
the size of each model so that each model has approximately the
same number of parameters. We intentionally made the models to be
small enough in order to avoid overfitting which can easily
distract the comparison.  This approach of comparing different
types of hidden units in neural networks has been done before,
for instance, by \citet{gulcehre2014learned}. See
Table~\ref{tab:models} for the details of the model
sizes.

\begin{table}[ht]
\centering
\begin{tabular}{c || c | c}
Unit & \# of Units   & \# of Parameters \\
\hline
\hline
\multicolumn{3}{c}{Polyphonic music modeling} \\
\hline
LSTM & 36 & $\approx 19.8 \times 10^3$ \\
GRU & 46 & $\approx 20.2 \times 10^3$ \\
$\tanh$ & 100 & $\approx 20.1 \times 10^3$ \\
\hline
\multicolumn{3}{c}{Speech signal modeling} \\
\hline
LSTM & 195 & $\approx 169.1 \times 10^3$ \\
GRU & 227 & $\approx 168.9 \times 10^3$ \\
$\tanh$ & 400 & $\approx 168.4 \times 10^3$ \\
\end{tabular}
\caption{The sizes of the models tested in the experiments.}
\label{tab:models}
\end{table}

\begin{table}[ht]
    \centering
    \begin{tabular}{ c | c | c || c | c | c }
        \hline
        \multicolumn{3}{c||}{} & $\tanh$ & GRU & LSTM \\
        \hline
        \hline
        \multirow{4}{*}{Music Datasets}
         & Nottingham &\specialcell{train \\ test}
                    &\specialcell{3.22   \\ \bf 3.13} &
                       \specialcell{2.79 \\ 3.23 } &
                       \specialcell{3.08 \\ 3.20 }  \\
        \cline{2-6}
        & JSB Chorales  &\specialcell{ train \\ test}
                        & \specialcell{8.82   \\ 9.10 } &
                        \specialcell{  6.94   \\ \bf 8.54 } &
                        \specialcell{  8.15   \\ 8.67 } \\
        \cline{2-6}
        & MuseData  &\specialcell{train \\ test}
                    & \specialcell{5.64   \\ 6.23 } &
                     \specialcell{ 5.06   \\ \bf 5.99 } &
                     \specialcell{ 5.18   \\ 6.23 } \\
        \cline{2-6}
        & Piano-midi  &\specialcell{train \\ test}
                    & \specialcell{5.64   \\ 9.03 } &
                     \specialcell{ 4.93   \\ \bf 8.82 } &
                     \specialcell{ 6.49   \\ 9.03 } \\
        \cline{1-6}
        \multirow{4}{*}{Ubisoft Datasets}
         & Ubisoft dataset A  &\specialcell{train \\ test}
                          &\specialcell{ 6.29 \\ 6.44 } &
                            \specialcell{2.31 \\ 3.59 } &
                            \specialcell{1.44 \\ \bf 2.70 } \\
        \cline{2-6}
         & Ubisoft dataset B  &\specialcell{train \\ test}
                           & \specialcell{7.61 \\ 7.62} &
                            \specialcell{0.38 \\ \bf 0.88} &
                            \specialcell{0.80 \\ 1.26} \\
        \hline
    \end{tabular}
    \caption{The average negative log-probabilities of the
        training and test sets.
    }
    \label{tab:model_perfs}
\end{table}

We train each model with RMSProp~\citep[see,
e.g.,][]{Hinton-Coursera2012} and use weight noise with standard
deviation fixed to $0.075$~\citep{graves2011practical}.  At every
update, we rescale the norm of the gradient to $1$, if it is
larger than $1$~\citep{Pascanu-et-al-ICML2013} to prevent
exploding gradients.  We select a learning rate (scalar
multiplier in RMSProp) to maximize the validation performance,
out of $10$ randomly chosen log-uniform candidates sampled from
$\mathcal{U}(-12, -6)~$\citep{bergstra2012random}.
The validation set is used for early-stop training as well.

\section{Results and Analysis}

Table~\ref{tab:model_perfs} lists all the results from our
experiments.
In the case of the polyphonic music datasets, the GRU-RNN
outperformed all the others (LSTM-RNN and $\tanh$-RNN) on all the
datasets except for the Nottingham. However, we can see that on
these music datasets, all the three models performed closely to
each other.

On the other hand, the RNNs with the gating units (GRU-RNN and
LSTM-RNN) clearly outperformed the more traditional $\tanh$-RNN
on both of the Ubisoft datasets. The LSTM-RNN was best with the
Ubisoft A, and with the Ubisoft B, the GRU-RNN performed best.

\begin{figure}[ht]
    \centering
    \begin{minipage}{1\textwidth}
        \centering
        Per epoch

        \begin{minipage}[b]{0.48\textwidth}
            \centering
            \includegraphics[width=1.\textwidth,clip=true,trim=0
            28 0 0]{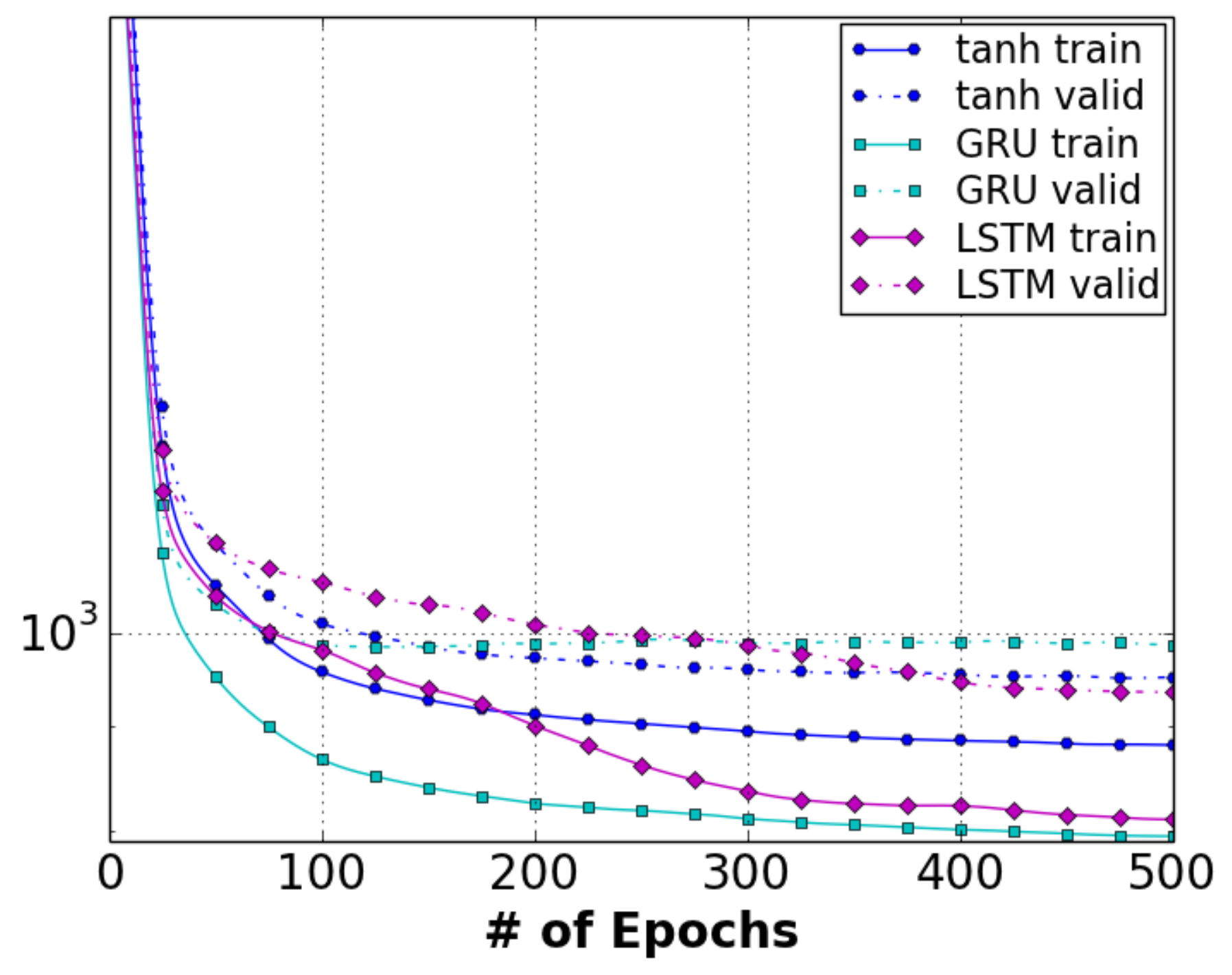}
        \end{minipage}
        \hfill
        \begin{minipage}[b]{0.48\textwidth}
            \centering
            \includegraphics[width=1.\textwidth,clip=true,trim=0
            28 0 0]{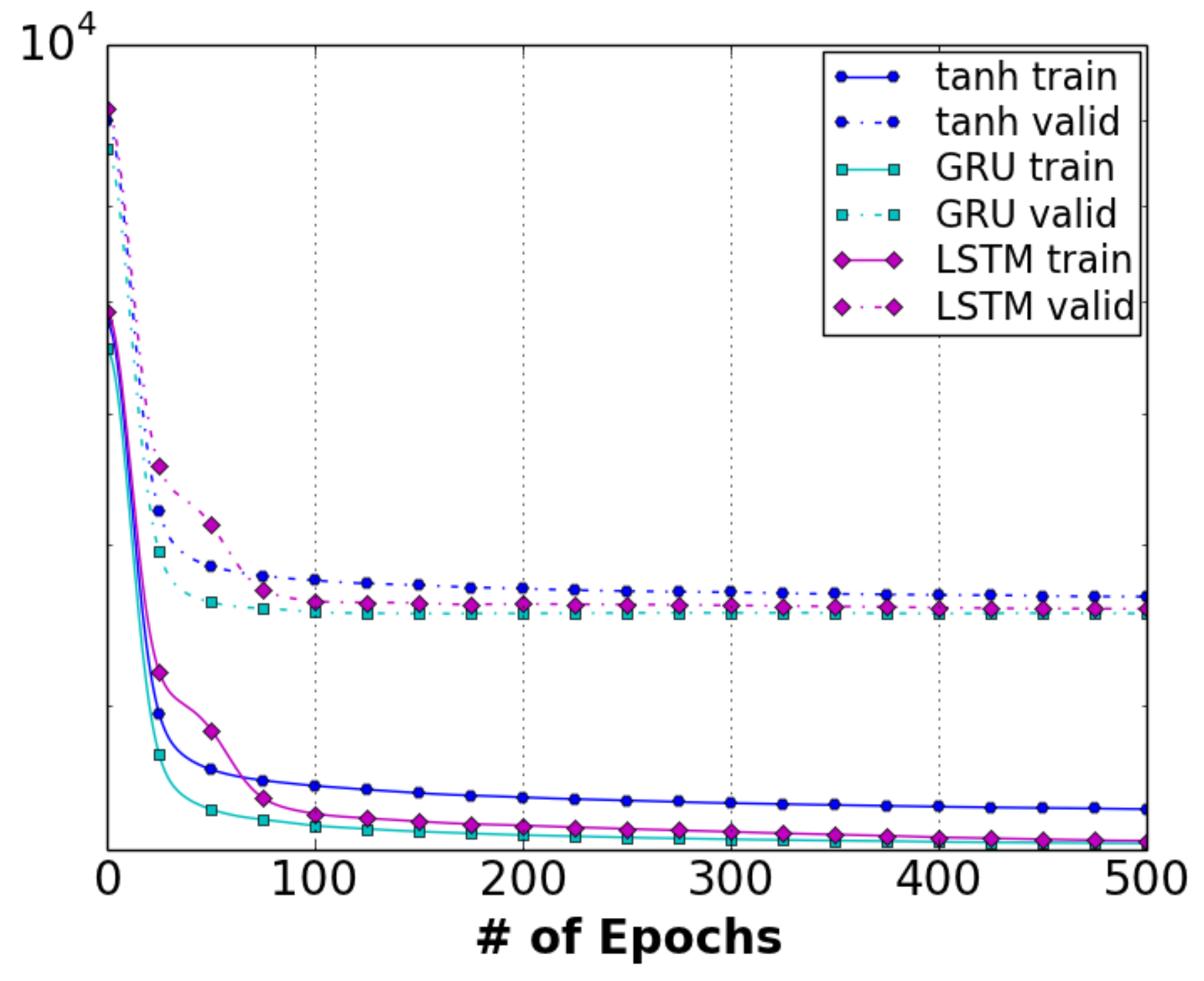}
        \end{minipage}

    \end{minipage}

    \vspace{4mm}
    \begin{minipage}{1\textwidth}
        \centering
        Wall Clock Time (seconds)

        \begin{minipage}[b]{0.48\textwidth}
            \centering
            \includegraphics[width=1.\textwidth,clip=true,trim=0 29 0 0]{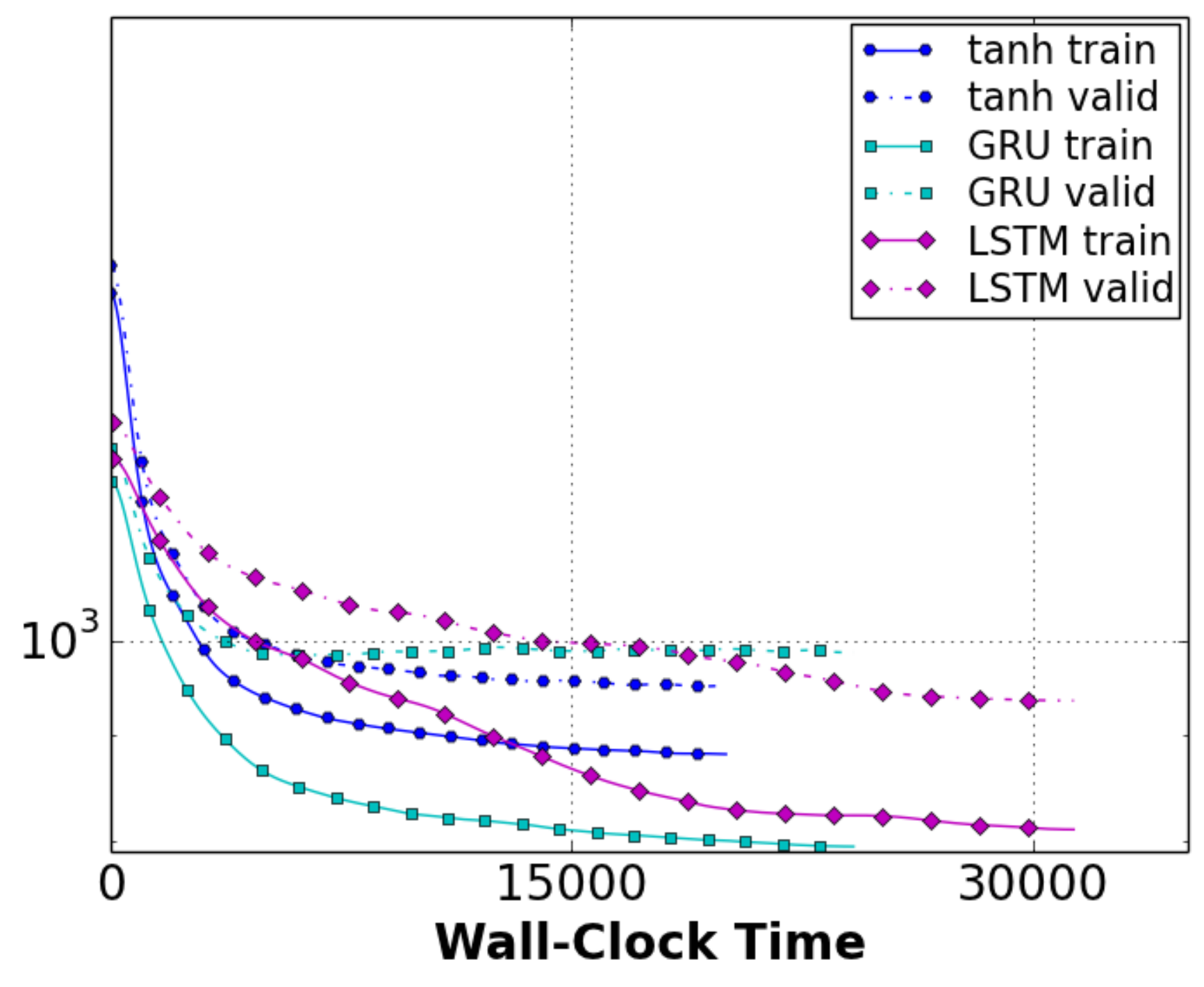}
        \end{minipage}
        \hfill
        \begin{minipage}[b]{0.48\textwidth}
            \centering
            \includegraphics[width=1.\textwidth,clip=true,trim=0 29 0 0]{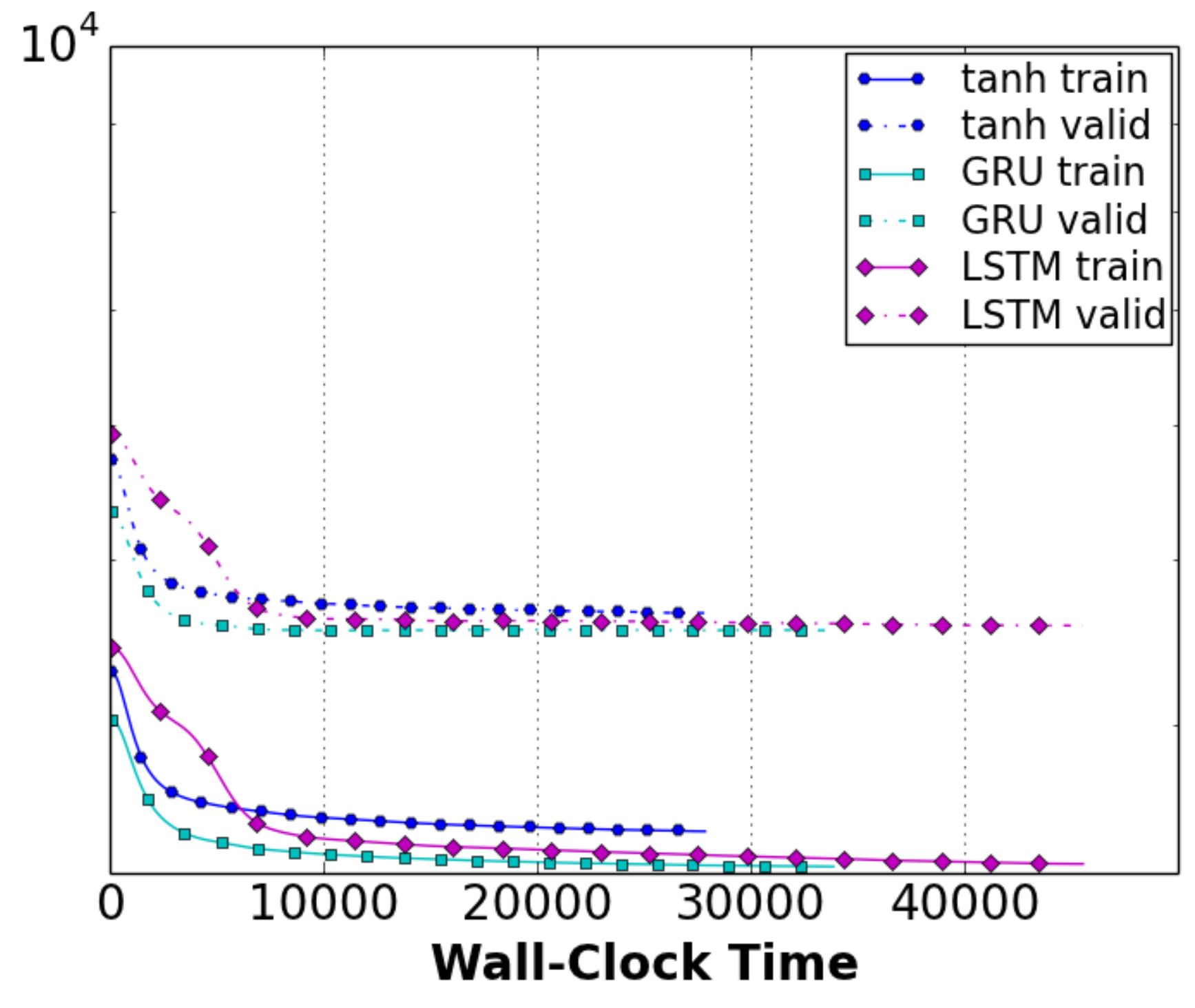}
        \end{minipage}

        \begin{minipage}{0.48\textwidth}
            \centering
            (a) Nottingham Dataset
        \end{minipage}
        \hfill
        \begin{minipage}{0.48\textwidth}
            \centering
            (b) MuseData Dataset
        \end{minipage}

    \end{minipage}

    \caption{Learning curves for training and validation sets of different
    types of units with respect to (top) the number of iterations and
    (bottom) the wall clock time. y-axis corresponds to the
negative-log likelihood of the model shown in log-scale.}
    \label{fig:music_results}
\end{figure}

\begin{figure}[ht]
    \centering
    \begin{minipage}{1\textwidth}
        \centering
        Per epoch

        \begin{minipage}[b]{0.48\textwidth}
            \centering
            \includegraphics[width=1.\textwidth,clip=true,trim=0 28 0 0]{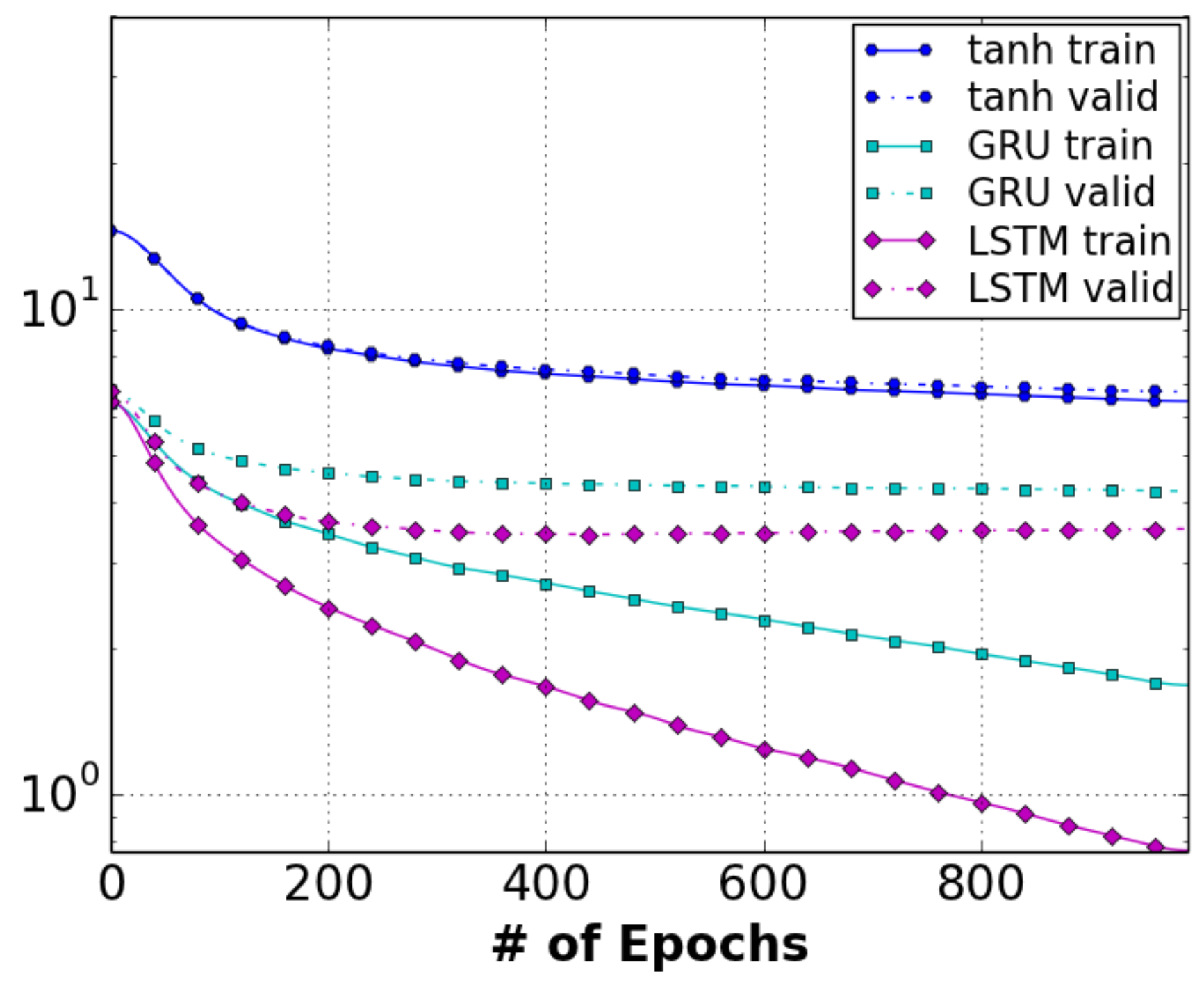}
        \end{minipage}
        \hfill
        \begin{minipage}[b]{0.48\textwidth}
            \centering
            \includegraphics[width=1.\textwidth,clip=true,trim=0 28 0 0]{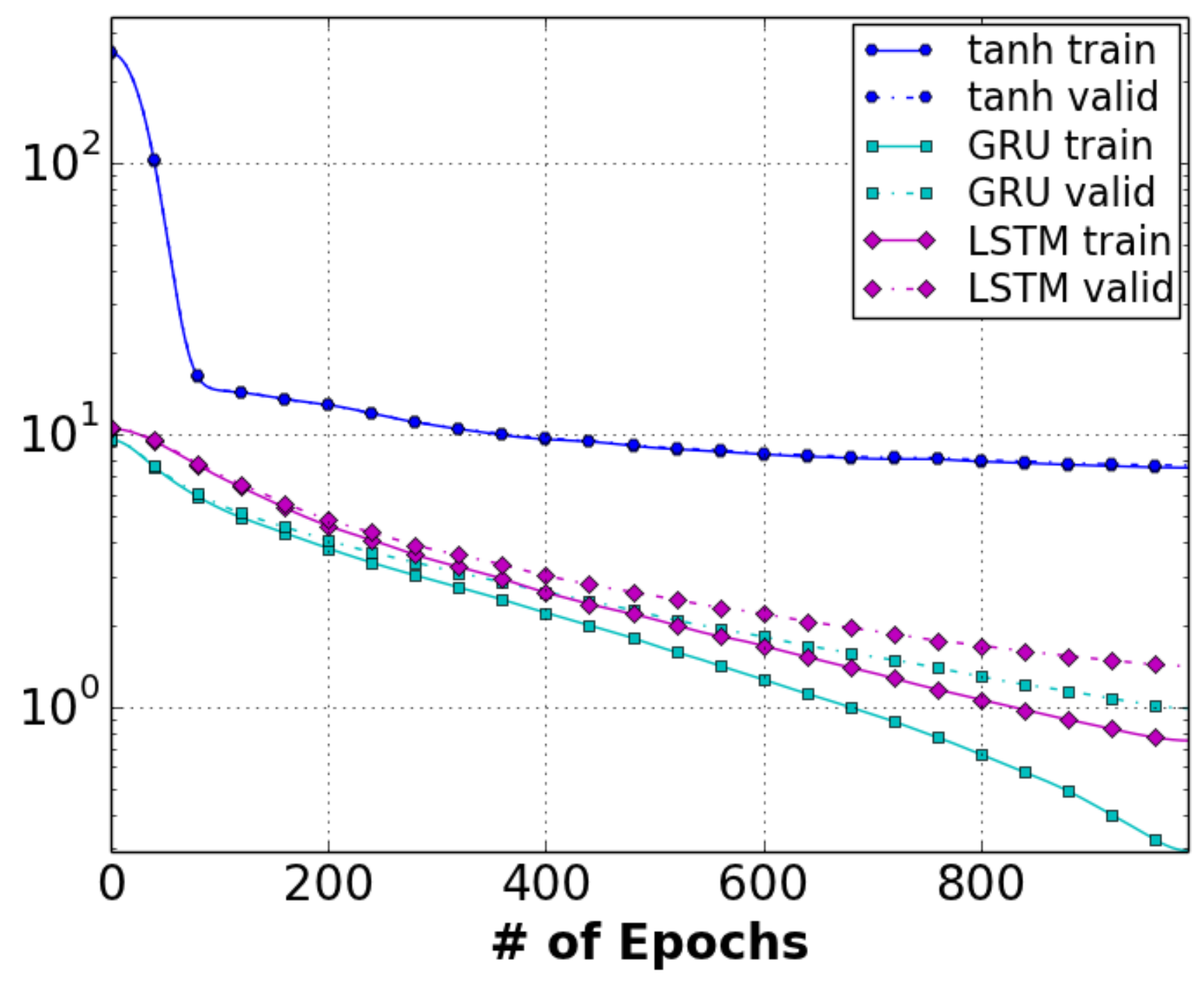}
        \end{minipage}

    \end{minipage}

    \vspace{4mm}
    \begin{minipage}{1\textwidth}
        \centering
        Wall Clock Time (seconds)

        \begin{minipage}[b]{0.48\textwidth}
            \centering
            \includegraphics[width=1.\textwidth,clip=true,trim=0 28 0 0]{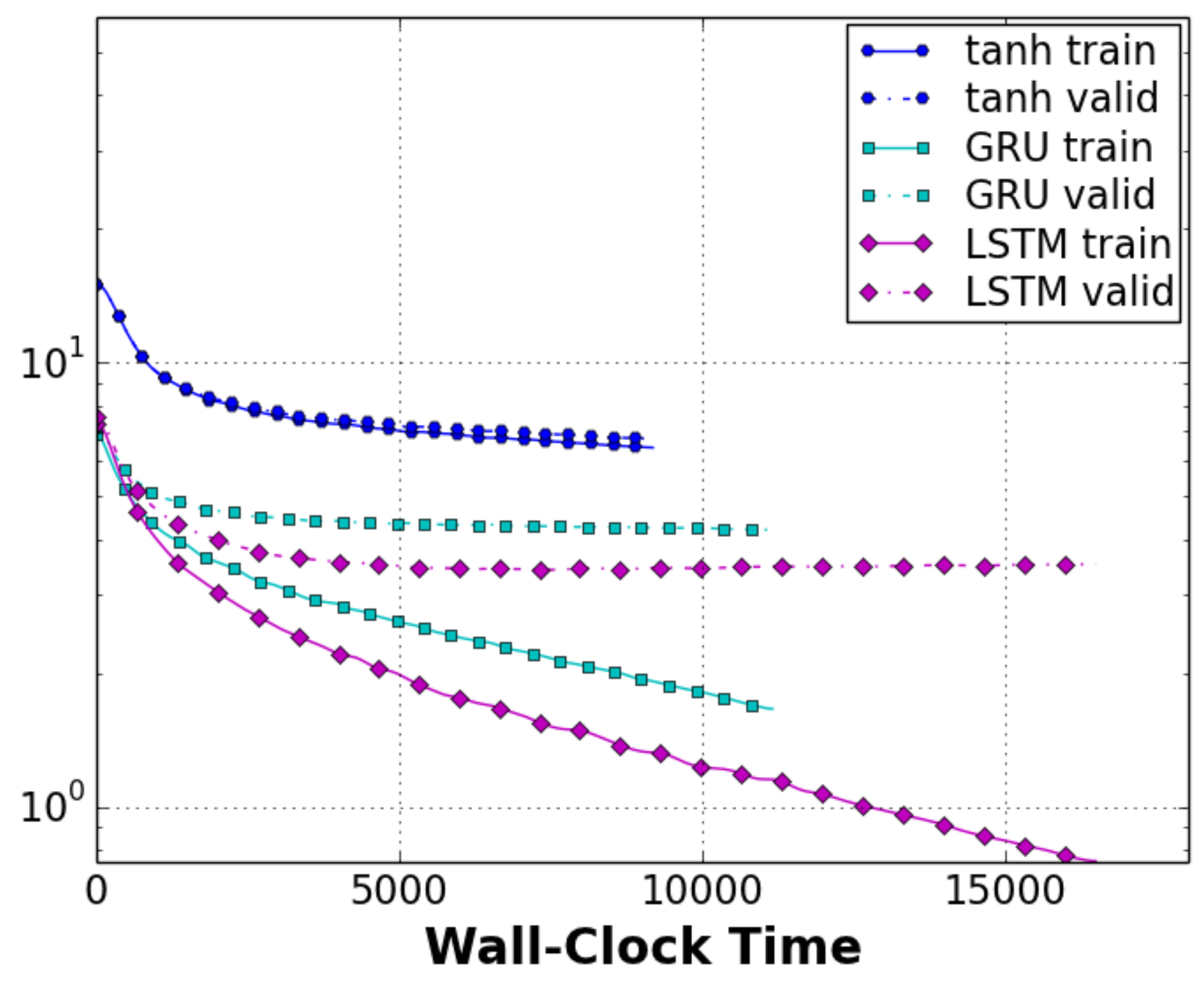}
        \end{minipage}
        \hfill
        \begin{minipage}[b]{0.48\textwidth}
            \centering
            \includegraphics[width=1.\textwidth,clip=true,trim=0 28 0 0]{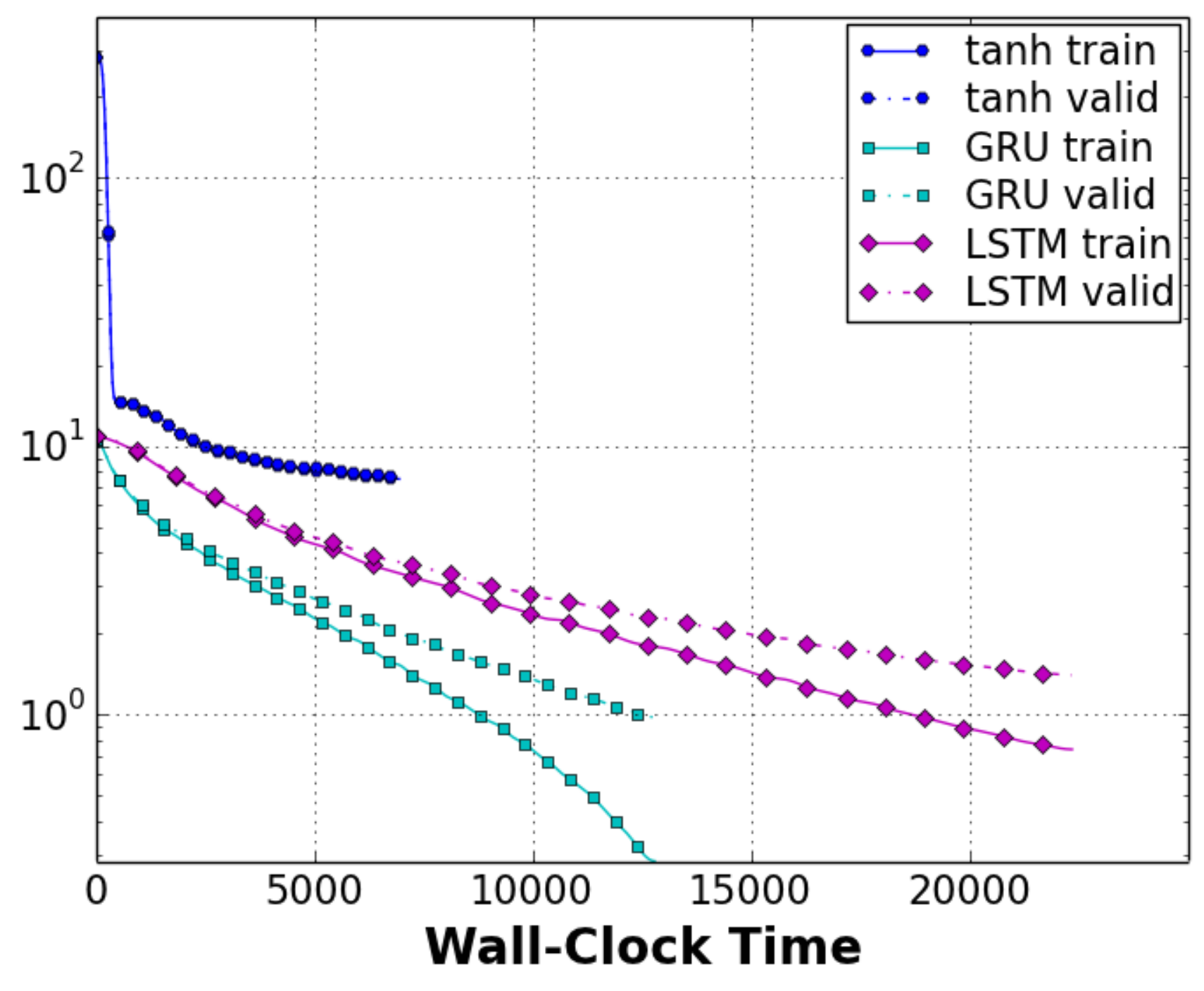}
        \end{minipage}

        \begin{minipage}{0.48\textwidth}
            \centering
            (a) Ubisoft Dataset A
        \end{minipage}
        \hfill
        \begin{minipage}{0.48\textwidth}
            \centering
            (b) Ubisoft Dataset B
        \end{minipage}

    \end{minipage}

    \caption{Learning curves for training and validation sets of different
        types of units with respect to (top) the number of iterations and
        (bottom) the wall clock time. x-axis is the number of epochs and
        y-axis corresponds to the negative-log likelihood of the model
        shown in log-scale.
    }
    \label{fig:ubi_results}
\end{figure}

In Figs.~\ref{fig:music_results}--\ref{fig:ubi_results}, we show
the learning curves of the best validation runs. In the case of
the music datasets (Fig.~\ref{fig:music_results}), we see that
the GRU-RNN makes faster progress in terms of both the number of
updates and actual CPU time. If we consider the Ubisoft datasets
(Fig.~\ref{fig:ubi_results}), it is clear that although the
computational requirement for each update in the $\tanh$-RNN is
much smaller than the other models, it did not make much
progress each update and eventually stopped making any progress
at much worse level.

These results clearly indicate the advantages of the gating units
over the more traditional recurrent units. Convergence is often
faster, and the final solutions tend to be better. However, our
results are not conclusive in comparing the LSTM and the GRU,
which suggests that the choice of the type of gated recurrent
unit may depend heavily on the dataset and corresponding task.

\section{Conclusion}

In this paper we empirically evaluated recurrent neural networks
(RNN) with three widely used recurrent units; (1) a traditional
$\tanh$ unit, (2) a long short-term memory (LSTM) unit and (3) a
recently proposed gated recurrent unit (GRU). Our evaluation
focused on the task of sequence modeling on a number of datasets
including polyphonic music data and raw speech signal data.

The evaluation clearly demonstrated the superiority of the gated
units; both the LSTM unit and GRU, over the traditional $\tanh$
unit. This was more evident with the more challenging task of raw
speech signal modeling. However, we could not make concrete
conclusion on which of the two gating units was better.

We consider the experiments in this paper as preliminary. In
order to understand better how a gated unit helps learning and to
separate out the contribution of each component, for instance
gating units in the LSTM unit or the GRU, of the gating units,
more thorough experiments will be required in the future.

\section*{Acknowledgments}

The authors would like to thank Ubisoft for providing the
datasets and for the support. The authors would like to thank
the developers of
Theano~\citep{bergstra+al:2010-scipy,Bastien-Theano-2012} and
Pylearn2~\citep{pylearn2_arxiv_2013}. We acknowledge the support
of the following agencies for research funding and computing
support: NSERC, Calcul Qu\'{e}bec, Compute Canada, the Canada
Research Chairs and CIFAR.

\newpage
\bibliography{strings,strings-shorter,ml,aigaion,myref}

\begin{thebibliography}{20}
\providecommand{\natexlab}[1]{#1}
\providecommand{\url}[1]{\texttt{#1}}
\expandafter\ifx\csname urlstyle\endcsname\relax
  \providecommand{\doi}[1]{doi: #1}\else
  \providecommand{\doi}{doi: \begingroup \urlstyle{rm}\Url}\fi

\bibitem[Bahdanau et~al.(2014)Bahdanau, Cho, and
  Bengio]{Bahdanau-et-al-arxiv2014}
D.~Bahdanau, K.~Cho, and Y.~Bengio.
\newblock Neural machine translation by jointly learning to align and
  translate.
\newblock Technical report, arXiv preprint arXiv:1409.0473, 2014.

\bibitem[Bastien et~al.(2012)Bastien, Lamblin, Pascanu, Bergstra, Goodfellow,
  Bergeron, Bouchard, and Bengio]{Bastien-Theano-2012}
F.~Bastien, P.~Lamblin, R.~Pascanu, J.~Bergstra, I.~J. Goodfellow, A.~Bergeron,
  N.~Bouchard, and Y.~Bengio.
\newblock Theano: new features and speed improvements.
\newblock Deep Learning and Unsupervised Feature Learning NIPS 2012 Workshop,
  2012.

\bibitem[Bengio et~al.(1994)Bengio, Simard, and Frasconi]{Bengio-trnn94}
Y.~Bengio, P.~Simard, and P.~Frasconi.
\newblock Learning long-term dependencies with gradient descent is difficult.
\newblock \emph{IEEE Transactions on Neural Networks}, 5\penalty0 (2):\penalty0
  157--166, 1994.

\bibitem[Bengio et~al.(2013)Bengio, Boulanger-Lewandowski, and
  Pascanu]{Bengio-et-al-ICASSP-2013}
Y.~Bengio, N.~Boulanger-Lewandowski, and R.~Pascanu.
\newblock Advances in optimizing recurrent networks.
\newblock In \emph{Proc. ICASSP 38}, 2013.

\bibitem[Bergstra and Bengio(2012)]{bergstra2012random}
J.~Bergstra and Y.~Bengio.
\newblock Random search for hyper-parameter optimization.
\newblock \emph{The Journal of Machine Learning Research}, 13\penalty0
  (1):\penalty0 281--305, 2012.

\bibitem[Bergstra et~al.(2010)Bergstra, Breuleux, Bastien, Lamblin, Pascanu,
  Desjardins, Turian, Warde-Farley, and Bengio]{bergstra+al:2010-scipy}
J.~Bergstra, O.~Breuleux, F.~Bastien, P.~Lamblin, R.~Pascanu, G.~Desjardins,
  J.~Turian, D.~Warde-Farley, and Y.~Bengio.
\newblock Theano: a {CPU} and {GPU} math expression compiler.
\newblock In \emph{Proceedings of the Python for Scientific Computing
  Conference ({SciPy})}, June 2010.
\newblock Oral Presentation.

\bibitem[Boulanger-Lewandowski et~al.(2012)Boulanger-Lewandowski, Bengio, and
  Vincent]{Boulanger-et-al-ICML2012}
N.~Boulanger-Lewandowski, Y.~Bengio, and P.~Vincent.
\newblock Modeling temporal dependencies in high-dimensional sequences:
  Application to polyphonic music generation and transcription.
\newblock In \emph{Proceedings of the Twenty-nine International Conference on
  Machine Learning (ICML'12)}. ACM, 2012.
\newblock URL \url{http://icml.cc/discuss/2012/590.html}.

\bibitem[Cho et~al.(2014)Cho, van Merrienboer, Bahdanau, and
  Bengio]{cho2014properties}
K.~Cho, B.~van Merrienboer, D.~Bahdanau, and Y.~Bengio.
\newblock On the properties of neural machine translation: Encoder-decoder
  approaches.
\newblock \emph{arXiv preprint arXiv:1409.1259}, 2014.

\bibitem[Goodfellow et~al.(2013)Goodfellow, Warde-Farley, Lamblin, Dumoulin,
  Mirza, Pascanu, Bergstra, Bastien, and Bengio]{pylearn2_arxiv_2013}
I.~J. Goodfellow, D.~Warde-Farley, P.~Lamblin, V.~Dumoulin, M.~Mirza,
  R.~Pascanu, J.~Bergstra, F.~Bastien, and Y.~Bengio.
\newblock Pylearn2: a machine learning research library.
\newblock \emph{arXiv preprint arXiv:1308.4214}, 2013.

\bibitem[Graves(2012)]{Graves-book2012}
A.~Graves.
\newblock \emph{Supervised Sequence Labelling with Recurrent Neural Networks}.
\newblock Studies in Computational Intelligence. Springer, 2012.

\bibitem[Graves(2011)]{graves2011practical}
A.~Graves.
\newblock Practical variational inference for neural networks.
\newblock In \emph{Advances in Neural Information Processing Systems}, pages
  2348--2356, 2011.

\bibitem[Graves(2013)]{graves2013generating}
A.~Graves.
\newblock Generating sequences with recurrent neural networks.
\newblock \emph{arXiv preprint arXiv:1308.0850}, 2013.

\bibitem[Graves et~al.(2013)Graves, Mohamed, and
  Hinton]{Graves-et-al-ICASSP2013}
A.~Graves, A.-r. Mohamed, and G.~Hinton.
\newblock Speech recognition with deep recurrent neural networks.
\newblock In \emph{ICASSP'2013}, pages 6645--6649. IEEE, 2013.

\bibitem[Gulcehre et~al.(2014)Gulcehre, Cho, Pascanu, and
  Bengio]{gulcehre2014learned}
C.~Gulcehre, K.~Cho, R.~Pascanu, and Y.~Bengio.
\newblock Learned-norm pooling for deep feedforward and recurrent neural
  networks.
\newblock In \emph{Machine Learning and Knowledge Discovery in Databases},
  pages 530--546. Springer, 2014.

\bibitem[Hinton(2012)]{Hinton-Coursera2012}
G.~Hinton.
\newblock Neural networks for machine learning.
\newblock Coursera, video lectures, 2012.

\bibitem[Hochreiter(1991)]{Hochreiter91}
S.~Hochreiter.
\newblock { Untersuchungen zu dynamischen neuronalen Netzen. Diploma thesis,
  Institut f\"{u}r Informatik, Lehrstuhl Prof. Brauer, Technische
  Universit\"{a}t M\"{u}nchen}, 1991.
\newblock URL \url{http://www7.informatik.tu-muenchen.de/~Ehochreit}.

\bibitem[Hochreiter and Schmidhuber(1997)]{Hochreiter+Schmidhuber-1997}
S.~Hochreiter and J.~Schmidhuber.
\newblock Long short-term memory.
\newblock \emph{Neural Computation}, 9\penalty0 (8):\penalty0 1735--1780, 1997.

\bibitem[Martens and Sutskever(2011)]{Martens+Sutskever-ICML2011}
J.~Martens and I.~Sutskever.
\newblock Learning recurrent neural networks with {H}essian-free optimization.
\newblock In \emph{Proc. ICML'2011}. ACM, 2011.

\bibitem[Pascanu et~al.(2013)Pascanu, Mikolov, and
  Bengio]{Pascanu-et-al-ICML2013}
R.~Pascanu, T.~Mikolov, and Y.~Bengio.
\newblock On the difficulty of training recurrent neural networks.
\newblock In \emph{Proceedings of the 30th International Conference on Machine
  Learning (ICML'13)}. ACM, 2013.
\newblock URL \url{http://icml.cc/2013/}.

\bibitem[Sutskever et~al.(2014)Sutskever, Vinyals, and
  Le]{Sutskever-et-al-arxiv2014}
I.~Sutskever, O.~Vinyals, and Q.~V. Le.
\newblock Sequence to sequence learning with neural networks.
\newblock Technical report, arXiv preprint arXiv:1409.3215, 2014.

\end{thebibliography}
\bibliographystyle{abbrvnat}

\end{document}